\documentclass[sigconf]{acmart}

\usepackage{subfigure}
\usepackage{balance}
\def\BibTeX{{\rm B\kern-.05em{\sc i\kern-.025em b}\kern-.08emT\kern-.1667em\lower.7ex\hbox{E}\kern-.125emX}}
    
\copyrightyear{2019}
\acmYear{2019}
\setcopyright{acmlicensed}

\begin{document}

%
\title{Subject Cross Validation in Human Activity Recognition}

%

\author{Akbar Dehghani}
\affiliation{%
 \institution{Concordia University}
 \city{Montreal}
 \country{Canada}
 }
 \email{a_ehg@encs.concordia.ca}
 
\author{Tristan Glatard}
\affiliation{%
  \institution{Concordia University}
  \city{Montreal}
  \country{Canada}
  }
  \email{tristan.glatard@concordia.ca}

\author{Emad Shihab}
\affiliation{%
  \institution{Concordia University}
  \city{Montreal}
  \country{Canada}
  }
\email{eshihab@cse.concordia.ca}

%
\begin{abstract}
K-fold Cross Validation is commonly used to evaluate classifiers
and tune their hyperparameters. However, it assumes that data points are Independent and Identically 
Distributed (i.i.d.) so that samples used in the training and test 
sets can be selected randomly and uniformly. In Human Activity Recognition datasets, 
we note that the samples produced by the same subjects are likely to be correlated due 
to diverse factors. Hence, 
k-fold cross validation may overestimate the performance of 
activity recognizers, in particular when overlapping sliding windows are used. 
In this paper, we investigate the effect of Subject 
Cross Validation on the performance of Human Activity Recognition, 
both with non-overlapping and with overlapping sliding windows. 
Results show that k-fold cross validation artificially 
increases the performance of recognizers by about 10\%, and even by 16\% when overlapping windows are used. In addition, we do not observe any performance gain from the use of overlapping windows. We conclude that Human Activity Recognition systems should be evaluated by Subject Cross Validation, and that overlapping windows are not worth their extra computational cost. 
\end{abstract}

%
%


%


%

%
\maketitle

\section{Introduction}

Wearable sensors and mobile devices keep transforming society at an 
increasing pace, creating a wide range of opportunities for knowledge 
extraction from new data sources. Human Activity Recognition (HAR), in 
particular, is an active research area due to its potential applications in 
security, virtual reality, sports training, and health care. For 
instance, it has been used to detect anomalous behaviors such as 
falls~\cite{bianchi2010barometric} and to track movement-related 
conditions in seniors~\cite{chen2014implementing}.

Most HAR systems rely on an Activity 
Recognition Process (ARP) to label activities. An ARP segments sensor data in
time windows, from which it extracts feature vectors that are then fed to a classifier. ARPs have several hyperparameters such as the size of the time windows and the 
type of features extracted. Works such as~\cite{banos2014window} 
and~\cite{sousa2017comparative} showed the impact of these parameters on 
ARP performance, and provided guidelines to optimize them.

The main technique to evaluate 
model performance and tune hyperparameters is k-fold Cross Validation (k-fold CV)~\cite{arlot2010survey}. It 
assumes that samples are Independent and Identically Distributed 
(i.i.d.), that is, data points are independently drawn from the same 
distribution. However, in HAR datasets, samples that belong to the same subject are likely to be related to each other, due to underlying environmental, biological and demographics factors. In addition, there is often a temporal correlation among samples 
of a subject, due to, for example, fatigue, training, experience. As a result, k-fold CV might overestimate classification performance by relying on correlations within 
subjects.

Furthermore, two types of sliding windows are used for time series segmentation: non-overlapping ones, in which time windows do not intersect, 
and overlapping ones, in which they do~\cite{lara2013survey}. Overlapping 
windows share data samples with each other, which further increases 
within-subject correlations~\cite{coggeshall2005asset}. This is another infringement to the i.i.d assumption in k-fold CV, potentially further overestimating
classification performance.    

To address this issue, this paper investigates the impact of  
Subject Cross Validation (Subject CV) on ARP, both with overlapping and 
with non-overlapping sliding windows. Through Subject CV, we quantify and discuss the performance overestimation resulting from k-fold CV. 

The question of validation is a timely topic in HAR. k-fold CV is 
widely employed, and examples are found both with overlapping and with 
non-overlapping windows. 
For instance, a well-known reference on HAR is 
~\cite{morris2014recofit}, which applies ARP on a dataset of 
accelerometer and gyroscope data collected from 114 participants over 
146 sessions. The authors address three major challenges namely (1) 
segmenting exercise from intermittent non-exercise periods, (2) 
recognizing which exercise is being performed, and (3) counting 
repetitions. Data points 
are windowed into 5-second overlapping windows sliding at 200ms and subsequently, 
each window is transformed into 224 features. Linear support vector 
machines (SVM) are used in the classification stage, and evaluated by 
k-fold CV. Spectacular performance is achieved, with precision and 
recall greater than 95\% to identify exercise periods, recognition 
of up to 99\% for circuits of 4 exercises, and counting  
accurate to $\pm$1 repetition 93\% of the time.

Another example is~\cite{banos2014window}, which we use as a baseline for our work. In their work, the authors evaluate several classifiers on 
a dataset of 17 subjects, using non-overlapping windows and k-fold CV. 
In particular, they investigate the selection of window sizes for 
optimal classification performance. They reach an F1-score close to 1 with the k-nearest neighbors classifier.
 
The main contributions of our work are:

\begin{itemize}

\item An in-depth investigation of how ARP performance is impacted by k-fold CV.

\item The proposal of Subject CV as a more reliable and robust validation of ARP.

\item A set of publicly available scripts to help the research community further shed light on the important topic of HAR validation.

\end{itemize}


In Section \ref{sec:background}, we present 
background in ARP, cross validation, and sliding windows. We also describe the baseline study from which we started. In Section 
\ref{sec:experiment setting} we explain our ARP setting, and in Section 
\ref{sec:result} we present our results. Finally, we discuss their impact in Section \ref{sec:discussion}, and 
we summarize our conclusions in Section \ref{sec:conclusion}.

\section{Background} \label{sec:background}
In this section, we overview the ARP process, describe the difference cross validation methods, and present the previous study  by Banos \emph{et al.}~\cite{banos2014window}, that we use as a baseline comparison.

\subsection{Activity Recognition Process}\label{subsec:ARP}

\begin{figure}[h]
    \centering
    \includegraphics[width=.5\textwidth]{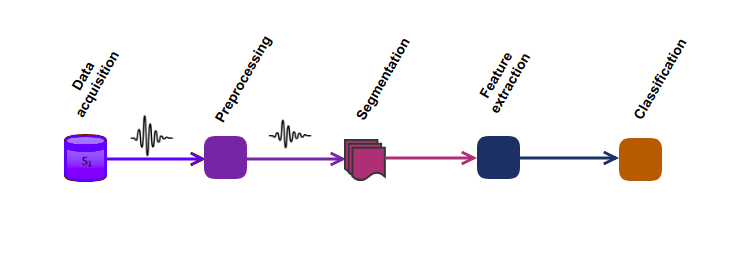}
    \caption{Human activity recognition process}
    \label{fig:tsprocess}
\end{figure}

ARP, also known as activity recognition 
chain, is composed of a sequence of signal processing, pattern recognition, and 
machine learning techniques~\cite{bulling2014tutorial}. It mainly 
consists of the 5 steps shown in Figure \ref{fig:tsprocess} and 
explained hereafter.

\begin{figure}[h]
    \centering
    \includegraphics[width=.4\textwidth]{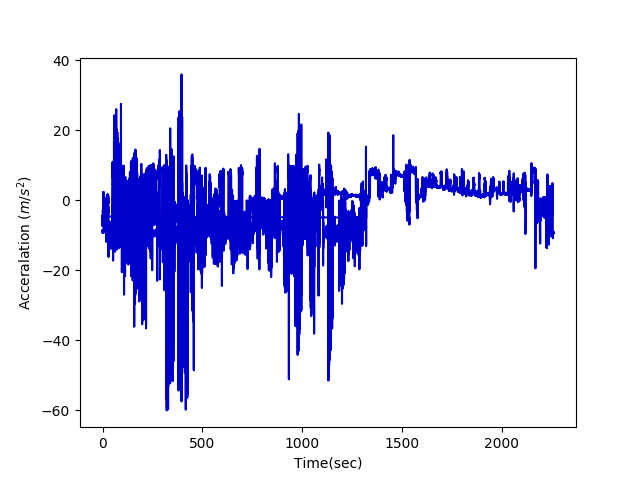}
    \caption{Example acceleration data extracted from~\cite{banos2012benchmark}.}
    \label{fig:signal}
\end{figure}

\noindent\textbf{Data acquisition.} Several sensors are attached to different body parts. They mostly acquire 3D acceleration, gyroscopic data and magnetic field measurements, as shown in Figure~\ref{fig:signal}. Sensor discretize signals at a given frequency, typically 50Hz for 
daily activities or 200Hz for fast sports, and transmit the resulting data points to 
the receiver. 

\noindent\textbf{Pre-processing.} Data points coming from sensors may
include artifacts of various origins such as 
electronic fluctuations, sensor malfunctions, and physical activities~\cite{arlot2010survey}. To eliminate such artifacts, filtering techniques are commonly applied, such as the Butterworth low-pass filter used in~\cite{morris2014recofit},~\cite{selles2005automated} and~\cite{najafi2003ambulatory}. 
Filtering should be used with care as it may also remove valuable information from the signals.


\noindent\textbf{Segmentation.}
Discrete data points produced by the sensors are partitioned into time 
windows labeled from the most frequent activity in the window. The number of data 
points in a time window, a.k.a the window size, heavily impacts the 
performance of the model~\cite{bulling2014tutorial}~\cite{banos2014window}. The current method 
to select the window size is empirical and time consuming: it simply tests different values to find the optimal one~\cite{bulling2014tutorial}.

\noindent\textbf{Feature extraction.}
Each time window is then transformed to a vector of features such as auto-correlation features~\cite{morris2014recofit}, or statistical 
moments. These features are then used to help discriminate various activities.

\noindent\textbf{Classification.}
Finally, a classifier is trained on the vector of features and corresponding 
labels, and assigns future 
observations to one of the learned activities. According to~\cite{lara2013survey}, Decision trees, Naive Bayes, SVM, k-nearest neighbors, Hidden Markov Models and ensemble classifiers such as Random Forest are the most important classifiers in HAR.

The window size in the segmentation step and the feature selection in feature extraction step are hyperparameters of the ARP, usually 
selected by trial and error as in~\cite{banos2014window}. They must 
be selected by the user and (as others have shown and we will show later in our paper) can greatly impact the performance of the model.

\subsection{Cross Validation Methods}
\label{sub:subjective CV}

The ultimate goal of ARP is to build a model that generalizes well, such that activities learned from a reduced set of subjects can be recognized in larger populations. Generalizability therefore has to be properly evaluated.

\begin{figure}[ht]
    \centering
    \includegraphics[width=.5\textwidth]{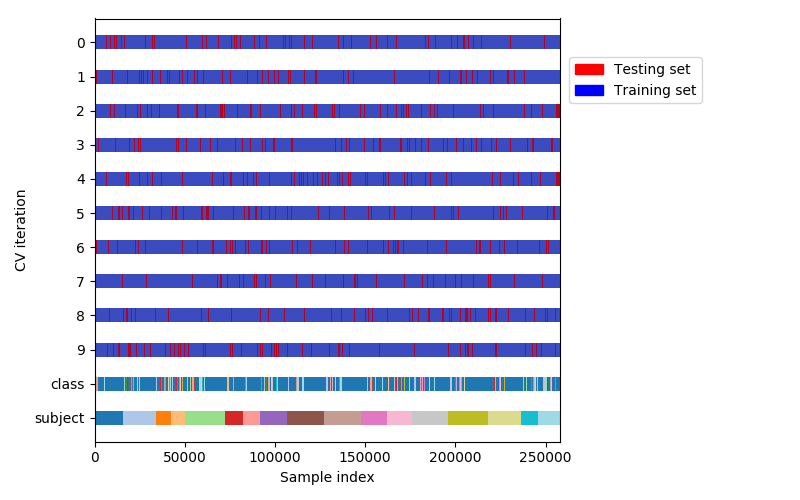}
    \caption{10-fold k-fold CV in overlapping windowed dataset (window size=0.5s).}
    \label{fig:Shuffle-cv}
\end{figure}

In k-fold CV (Figure~\ref{fig:Shuffle-cv}) the overall data is randomly partitioned in k equal subsets. The model is then trained on k-1 subsets, and the remaining one is used for testing~\cite{trevor2009elements}. The main assumption of k-fold CV is that samples are Independent and Identically Distributed (i.i.d.)~\cite{arlot2010survey}, which means that all the data points are sampled independently from the same distribution. Under such an assumption, the test set can be any part of the dataset.

However, samples drawn from a given subject are likely to \emph{not} be independent, for two reasons. First, there is a strong inter-subject variability in the way activities are conducted~\cite{bulling2014tutorial}. This means that the similarity of samples drawn from the same subject is likely to be higher than that of samples drawn from different subjects. Several factors might explain such variability, including sex, gender, age or experience. Second, there is a temporal dependence between activities performed by the same subject: the similarity between samples drawn in a short time interval, for instance in the same training session in case of training activities, will most likely be higher than that of samples drawn further apart in time. This is due to factors such as fatigue and training.

\begin{figure}[h]
    \centering
    \includegraphics[width=.5\textwidth]{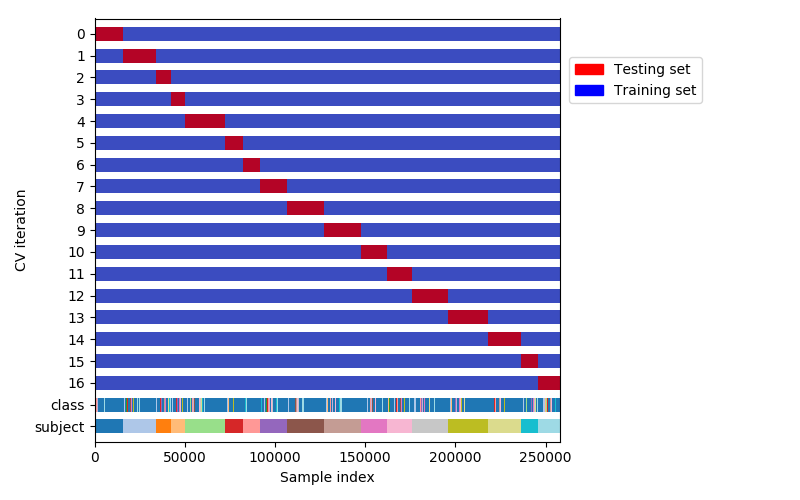}
    \caption{17-fold Subject CV process in overlapping windowed dataset (window size=0.5s).}
    \label{fig:Subjective-cv}
\end{figure}

To address these issues, Subject CV (Figure \ref{fig:Subjective-cv}) splits the training and testing sets by subject. That is, in each fold the model is trained on all the subjects except one, which is used for testing. The intra-subject dependencies present in k-fold CV are hence removed. In this case, the number of folds is lower or equal to the number of subjects in the dataset.

\begin{figure}[htp]
  \centering
  \subfigure[Non-overlapping]{
\label{subfig:NOSW}\includegraphics[scale=0.3]{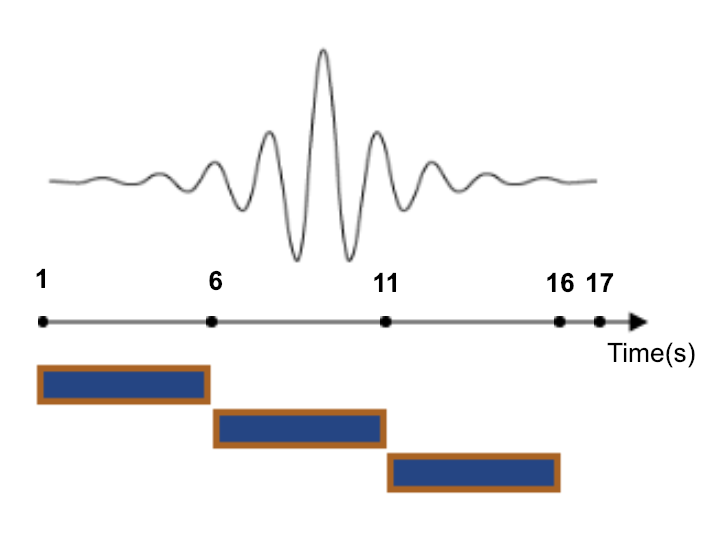}}\quad
  \subfigure[Overlapping-2 s sharing ]{\label{subfig:OSW}\includegraphics[scale=0.3]{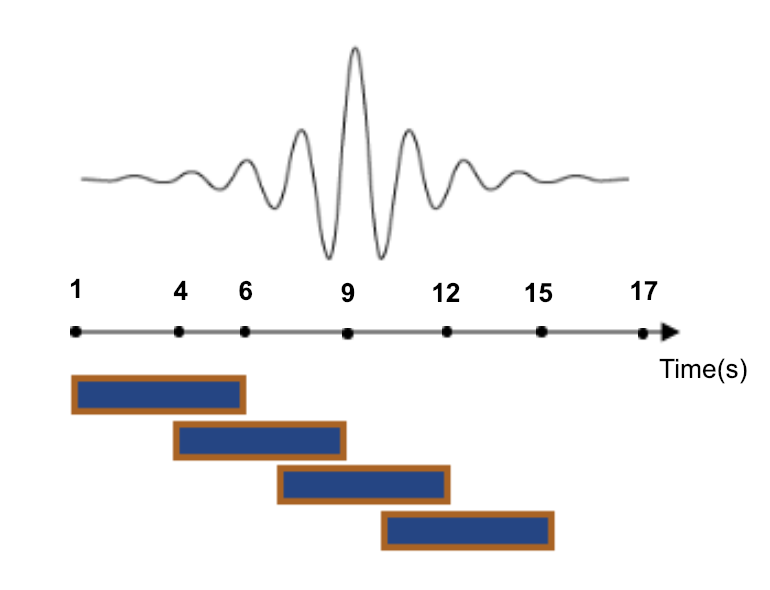}}

   \caption{5-second sliding windows. }
   \label{fig:SlidingWindow}
\end{figure}

\noindent\textbf{Overlapping sliding windows.} In the segmentation step of ARP, the data points are windowed to capture the dynamics of activities. This process assumes that each 
window is an approximation of the signal for which the classifier will 
have to make a decision. Figure~\ref{fig:SlidingWindow} illustrates the 
non-overlapping and overlapping windowing techniques. 
Non-overlapping windows are commonly considered less accurate compared to
overlapping windows because (1) overlapping windows result in more data 
points, which usually increases the performance of classifiers, and (2)
non-overlapping windows may miss important events in the dataset. 
A more formal discussion of the superiority of overlapping windows over non-overlapping ones
can be found in~\cite{coggeshall2005asset}.


\subsection{Baseline study} \label{sub:theirwork}

\begin{figure}[htp]
    \centering
    \includegraphics[width=.5\textwidth]{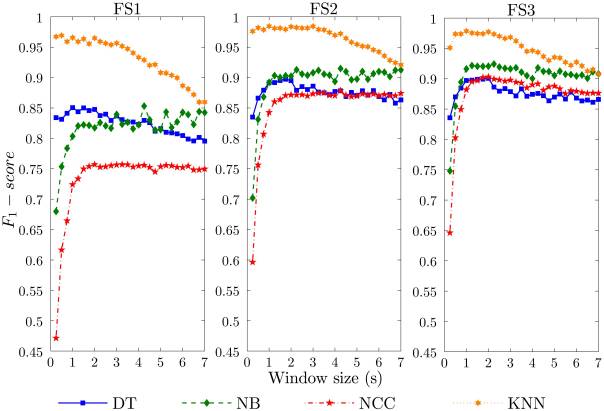}
    \caption{Impact of window size on F1-score of HAR system. Reproduced from~\cite{banos2014window}.}
    \label{fig:Banos et al result}
\end{figure}

Our study reproduces and extends the work in~\cite{banos2014window}, where the authors apply ARP on the 17-subject dataset
described in~\cite{banos2012benchmark}. They segment the signals without pre-processing, 
using non-overlapping time windows with diverse sizes ranging from 
0.25s to 7s. They consider three different feature sets, namely the
mean only (FS1), the mean and standard deviation (FS2), and the mean, 
standard deviation, maximum, minimum and mean crossing rate (FS3). They 
compare four classifiers: C4.5 Decision Trees (DT), K-Nearest Neighbors 
(KNN) with k=3, Naive Bayes (NB) and Nearest Centroid Classifier (NCC). 
They report F1-scores through a ten-fold cross-validation. 
Finally, they select optimal window sizes through a grid search process 
over specified values. Figure~\ref{fig:Banos et al result} illustrates their results for different feature sets and classifiers.

\section{Experiment design} \label{sec:experiment setting}
The motivation for our study is to investigate the effect of Subject CV on ARP. We consider both overlapping and non-overlapping sliding windows. 

\subsection{Dataset} \label{sec:dataset} 
 We use the dataset described in~\cite{banos2012benchmark}, one of the most complete public datasets for HAR in terms of the number of activities and subjects. The dataset consists of data collected from 17 subjects of diverse profiles while wearing 9 Xsens\footnote{\url{https://www.xsens.com}} inertial measurement units on different parts of their body. Subjects performed 33 fitness activities (Table \ref{tab:Activites}) ranging from warm up to fitness exercises in an out-of-lab environment. Each sensor provides tri-directional acceleration, gyroscope, and magnetic field measurements as well as orientation estimates in quaternion format (4D). Only acceleration was used in~\cite{banos2014window}, and hence in our study. The dataset also provides data for three sensor displacement scenarios namely ``default", ``self-placement" and ``mutual-displacement" to compare the sensor anomalies, but as in~\cite{banos2014window}, only the data from default scenario is used in our study. 

\begin{table*}
  \centering
\begin{tabular}{|c|c|}
\hline 
\multicolumn{2}{|c|}{Activities}\tabularnewline
\hline 
\hline 
Walking (1 min) & Upper trunk and lower body opposite twist (20x)\tabularnewline
\hline 
Jogging (1 min) & Arms lateral elevation (20x)\tabularnewline
\hline 
Running (1 min) & Arms frontal elevation (20x)\tabularnewline
\hline 
Jump up (20x) & Frontal hand claps (20x)\tabularnewline
\hline 
Jump front \& back (20x) & Arms frontal crossing (20x)\tabularnewline
\hline 
Jump sideways (20x) & Shoulders high amplitude rotation (20x)\tabularnewline
\hline 
Jump leg/arms open/closed (20x) & Shoulders low amplitude rotation (20x)\tabularnewline
\hline 
Jump rope (20x) & Arms inner rotation (20x)\tabularnewline
\hline 
Trunk twist (arms outstretched) (20x) & Knees (alternatively) to the breast (20x)\tabularnewline
\hline 
Trunk twist (elbows bended) (20x) & Heels (alternatively) to the backside (20x)\tabularnewline
\hline 
Waist bends forward (20x) & Knees bending (crouching) (20x)\tabularnewline
\hline 
Waist rotation (20x) & Knees (alternatively) bend forward (20x)\tabularnewline
\hline 
Waist bends (reach foot with opposite hand) (20x) & Rotation on the knees (20x)\tabularnewline
\hline 
Reach heels backwards (20x) & Rowing (1 min)\tabularnewline
\hline 
Lateral bend (10x to the left + 10x to the right) & Elliptic bike (1 min)\tabularnewline
\hline 
Lateral bend arm up (10x to the left + 10x to the right) & Cycling (1 min)\tabularnewline
\hline 
Repetitive forward stretching (20x) & \tabularnewline
\hline 
\end{tabular}
        \caption{Activity set in the dataset.}        
        \label{tab:Activites}
\end{table*} 


\subsection{Study setup} 

 Similar to~\cite{banos2014window}, we did not apply any pre-processing to the dataset. We used both overlapping and non-overlapping windows. Overlapping windows were sliding at 200~ms, with window sizes ranging from 0.25~s to 7~s. For instance, a 5-second window shared 4.8~s of data with the previous one. For non-overlapping windows, we used the same settings as in~\cite{banos2014window}: disjoint windows with sizes ranging from 0.25~s to 7~s. We used the same feature sets as in~\cite{banos2014window}, namely FS1 (mean only), FS2 (mean and standard deviation) and FS3 (mean, standard deviation, maximum, minimum and mean crossing rate). Finally, for the classification part, we used the following classifiers: Decision Tree (DT), K-nearest neighbors (KNN, K=3), Naive Bayes (NB), Nearest Centroid Classifier (NCC). We used these classifiers as implemented in scikit-learn 0.20~\cite{pedregosa2011scikit}.
 
 To evaluate model performance, we used both k-fold CV, as in~\cite{banos2014window}, and Subject CV. We use the F1-score as performance measure, computed as follows:
 \begin{center}
      $F1= \frac{2\times(precision \times recall)}{(precision + recall)}$
 \end{center}
In our multi-class scenario, we compute the F1-score using the total number of true positives, false negatives, and false positives across all the classes. This is known as f1\_micro in scikit-learn~\cite{pedregosa2011scikit}, by opposition with f1\_macro that averages F1-scores computed in each class individually.


\section{Results} 
\label{sec:result}

The goal of our study is to explore the importance of Subject CV in ARP, both with non-overlapping and with overlapping windows. In this section, the impact of Subject CV and overlapping windows on the system described in~\cite{banos2014window} is explored.

\subsection{Experiment 1: k-fold CV, non-overlapping windows} \label{reproduce}

    In this experiment, we intend to reproduce the work in~\cite{banos2014window} and use it as a baseline
    for further evaluations. 
    We applied the ARP as explained in Section~\ref{sub:theirwork}, on the dataset described in Section~\ref{sec:dataset}, using the classifiers in~\cite{banos2014window}. For each window size, we partitioned the dataset in non-overlapping windows and extracted feature sets FS1, FS2 and FS3 in each window. We trained the classifiers on the resulting feature vectors, and measured their average F1-score over 10-fold CV.
    
    Figure~\ref{fig:NO-iid-results} shows the F1-scores of the classifiers for different window sizes. The classifiers can be categorized in two performance groups: (1) KNN and DT show good performance (average F1-score: 0.8969 for FS3), while (2) NB and NCC show poor performance (average F1-score: 0.6578 for FS3). Similar to~\cite{banos2014window} (Figure~\ref{fig:Banos et al result}), the performance of the first group (KNN and DT) decreased with the size of the window. For the second group, the F1-score increased until it reached a maximum, around 1 second, then decreased (NCC) or remained almost constant (NB). Quantitatively, the F1-scores of KNN are very similar to the ones in~\cite{banos2014window}; KNN ranked first among all classifiers for all feature sets, with F1-scores above 0.96. However, the F1-scores of NB, DT and NCC are slightly different from the scores in \cite{banos2014window}. For FS3, the average F1-scores were 0.89 (DT), 0.71 (NB) and 0.6 (NCC) while in~\cite{banos2014window}, we estimate them as 0.86, 0.91 and 0.88 respectively. As in~\cite{banos2014window}, the performance of the models improves on average as the feature set becomes richer, i.e., from FS1 to FS2, and from FS2 to FS3. In particular, FS2 notably improves the performance of NCC and NB compared to FS1. Finally, as in~\cite{banos2014window}, KNN and DT perform best for the smallest window size (0.25~s), but NB and NCC need larger window sizes. In both case, 1~s was a cut-off value for the window size, for all feature sets and all classifiers: beyond this value, no important performance benefits were obtained.

    \subsection{Experiment 2: k-fold CV, overlapping windows} \label{subsec:classic-O}
   
    As explained earlier, overlapping windows increase the dependence between time windows, which potentially increases classification performance measured through k-fold CV. In this experiment, we use overlapping windows to explore their effect on the ARP system described in~\cite{banos2014window}.
    
     The only difference between this experiment and Experiment 1 is in the segmentation step. Here, each window slides at 200~ms. We selected 200~ms as in~\cite{morris2014recofit}.
     
     Our results are shown in Figure~\ref{fig:O-iid-results}. Similar to  Experiment 1, we observe the same two performance groups. However, the trend is inverted in the first group: now, the F1-score increases with the window size. Quantitatively, the F1-score of KNN and DT increased by about 10\% on average, but that of NB and NCC remained similar. KNN was the best performing classifier for all feature sets, as in the Experiment 1. However, KNN no longer allows us to maximally reduce the window size. As in the Experiment 1, overall, the richer the feature set used, the higher the performance obtained. Optimal window sizes (peak values in the Figures) are very different compared to the Experiment 1, in particular for KNN and DT.
     
     These results show that overlapping sliding windows further overestimate classification performance by k-fold CV, due to the dependency between datapoints.

\subsection{Experiment 3: Subject CV, non-overlapping windows}

    In this experiment we measured the effect of Subject CV on the ARP system of~\cite{banos2014window}. 
    The only difference between this experiment and Experiment 1 is the evaluation approach. We applied the ARP with the same settings as previously, except that we replaced k-fold CV by Subject CV.
    
\begin{figure*}[htp]
  \centering
  \subfigure[FS1]{
\label{fig:NO-FS1-iid}\includegraphics[scale=0.30]{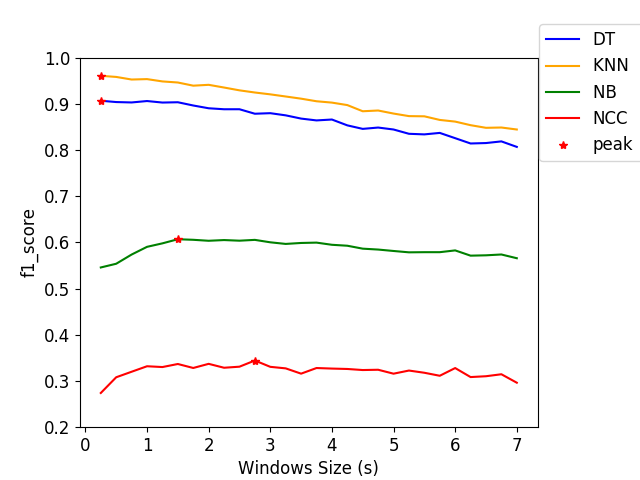}}\quad
  \subfigure[FS2]{\label{fig:NO-FS2-iid}\includegraphics[scale=0.30]{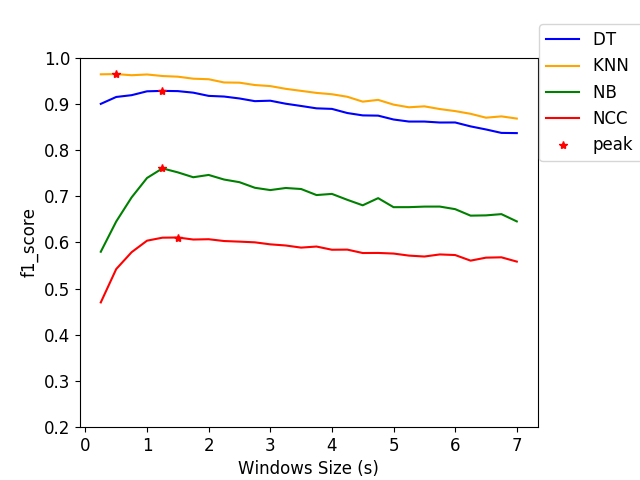}}
  \subfigure[FS3]{\label{fig:NO-FS3-iid}\includegraphics[scale=0.30]{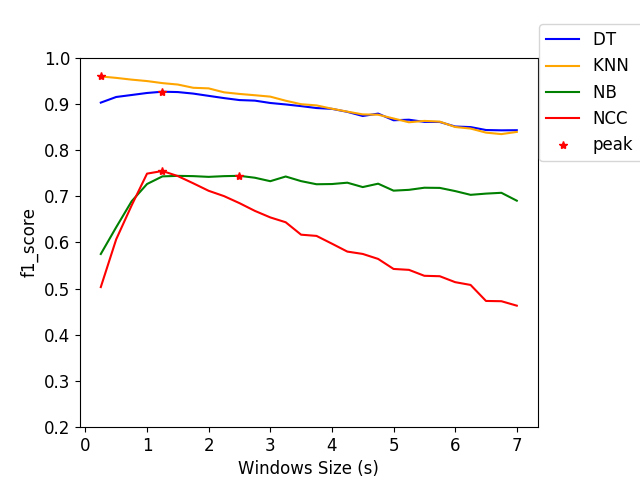}}
   
   \caption{Experiment 1 -- Non-overlapping windowing- k-fold CV process}
   \label{fig:NO-iid-results}
\end{figure*}

\begin{figure*}[htp]
  \centering
  \subfigure[FS1]{
\label{fig:NO-FS1-sbj}\includegraphics[scale=0.30]{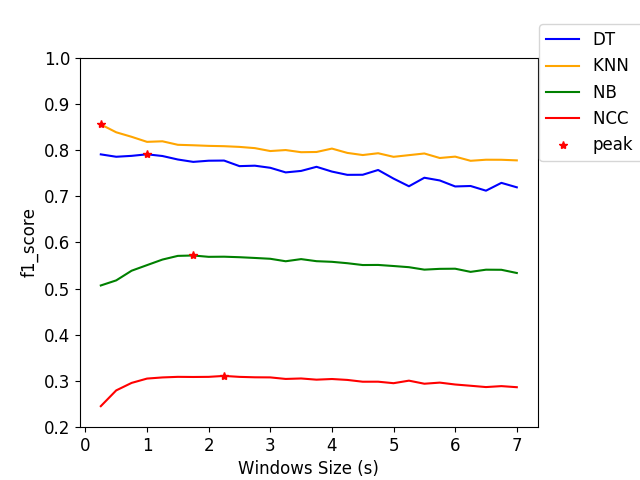}}\quad
  \subfigure[FS2]{\label{fig:NO-FS2-sbj}\includegraphics[scale=0.30]{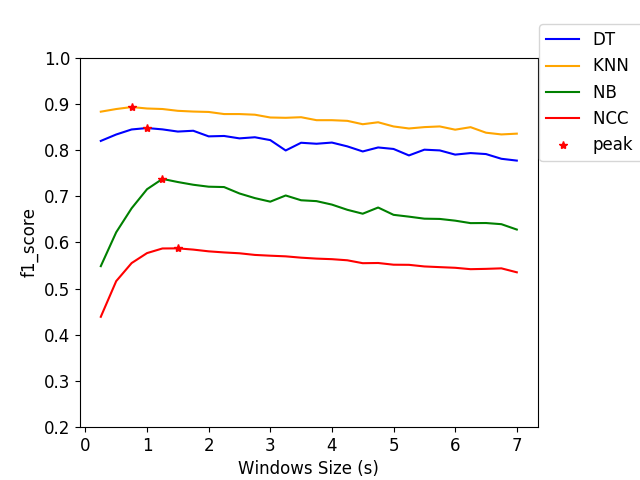}}
  \subfigure[FS3]{\label{fig:NO-FS3-sbj}\includegraphics[scale=0.30]{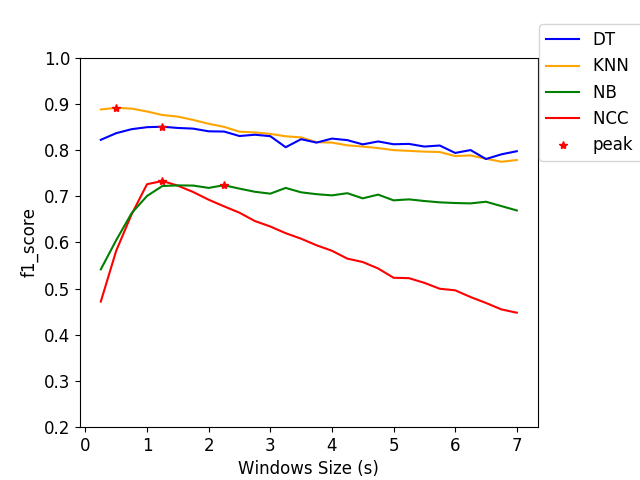}}
   
   \caption{Experiment 3 -- Non-overlapping windowing- Subject CV process}
    \label{fig:NO-subjective-results}

\end{figure*}

    Our results are shown in Figure~\ref{fig:NO-subjective-results}. The two groups of classifiers remained unchanged from Figure~\ref{fig:NO-iid-results}. Although the trends are similar, important quantitative differences between F1-scores are observed for KNN and DT: here, F1-scores of KNN and DT peaked at 0.89 and 0.85, while in Figure~\ref{fig:Banos et al result}, they reached their peaks at 0.99 and 0.93 respectively. In comparison, the performance for NB and NCC remained stable. As before, the F1-score of classifiers increased as the feature sets enriched. Regarding the optimal window sizes, in comparison to those in~\cite{banos2014window}, they remained almost unchanged and 1 second remained a cut-off value.
    
    Overall the results show that Subject CV removes the overestimation observed in k-fold CV.

\subsection{Experiment 4: Subject CV, overlapping sliding windows}

In Experiment  3, we observed that applying Subject CV instead of k-fold CV heavily changes the performance of the best classifiers. Here we measure how overlapping windows impact the results of Experiment  3. 

To conduct this experiment, the parameters used in Experiment  3 remained the same, except the sliding windows. Here, we use overlapping sliding windows with 200~ms sliding.

\begin{figure*}[htp]
  \centering
  \subfigure[FS1]{
\label{fig:O-FS1-sbj}\includegraphics[scale=0.30]{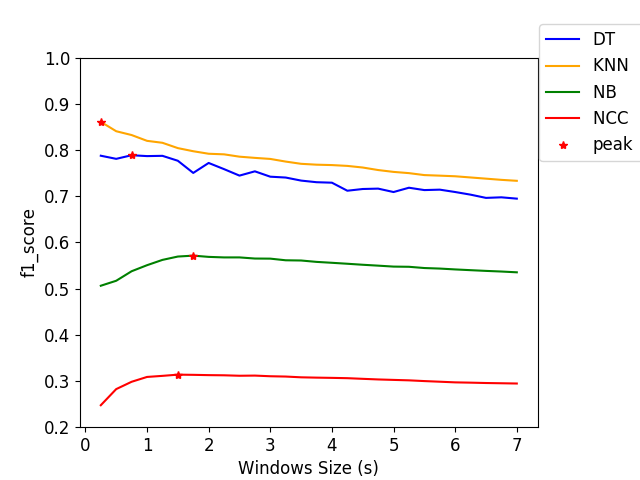}}\quad
  \subfigure[FS2]{\label{fig:O-FS2-sbj}\includegraphics[scale=0.30]{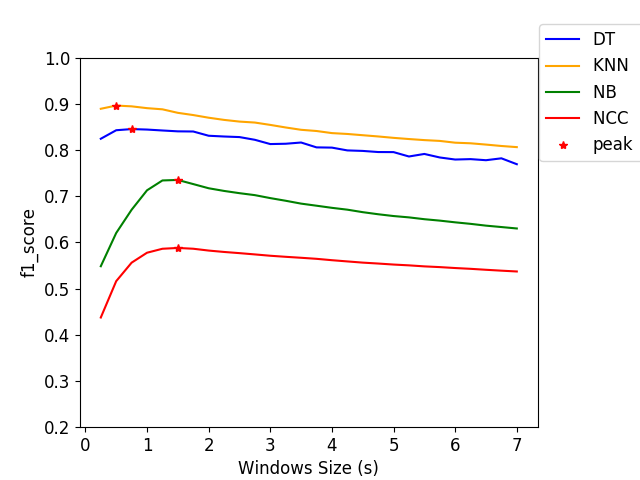}}
  \subfigure[FS3]{\label{fig:O-FS3-sbj}\includegraphics[scale=0.30]{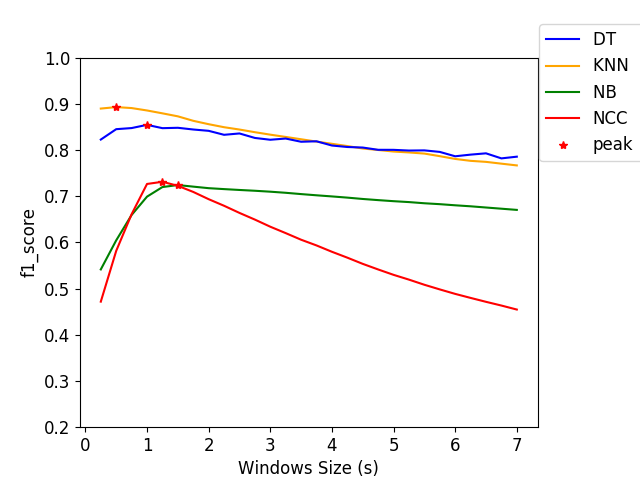}}
   
   \caption{Experiment 4 -- Overlapping windowing- Subject CV process}
    \label{fig:O-subjective-results}

\end{figure*}

Our results are reported in Figure~\ref{fig:O-subjective-results}. As in the previous experiments, the classifiers can be categorized in two groups. The relationship between window size and F1-score stayed unchanged compared to Experiment  3 and conducted experiment in Experiment  1. More precisely, contrary to the analyses described in Section \ref{subsec:classic-O}, using overlapping sliding windows did not inverse the trend of KNN and DT. This finding shows that such an approach is not a determining factor when used with Subject CV process. Quantitatively, the F1-score of models was the same as in Experiment 3. Likewise, using richer feature sets led to better performance. As for optimal window size, in general, there were no noticeable differences compared to Experiment  3.   

Overall the results of this experiment show that the artificial performance increase of overlapping windows observed in k-fold CV (Experiment 3) does not appear with Subject CV.

\begin{figure*}[htp]
  \centering
  \subfigure[FS1]{
\label{fig:O-FS1-iid}\includegraphics[scale=0.30]{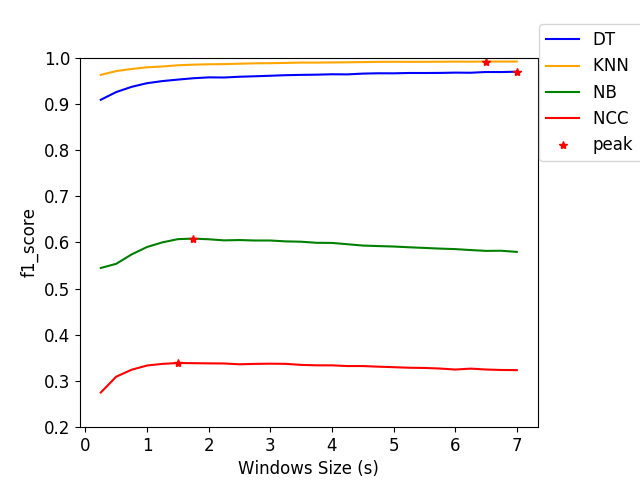}}\quad
  \subfigure[FS2]{\label{fig:O-FS2-iid}\includegraphics[scale=0.30]{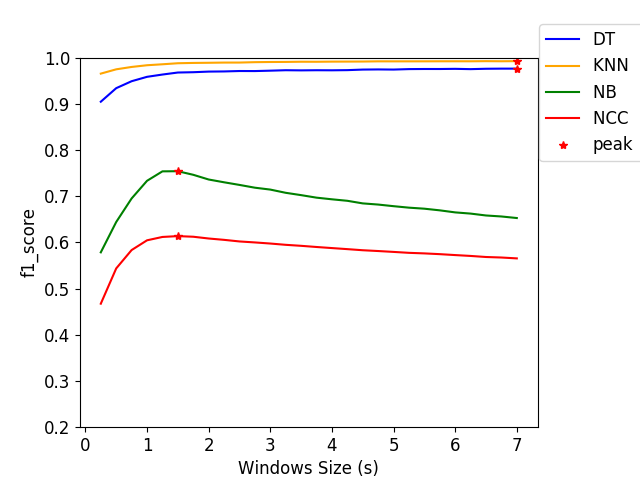}}
  \subfigure[FS3]{\label{fig:O-FS3-iid}\includegraphics[scale=0.30]{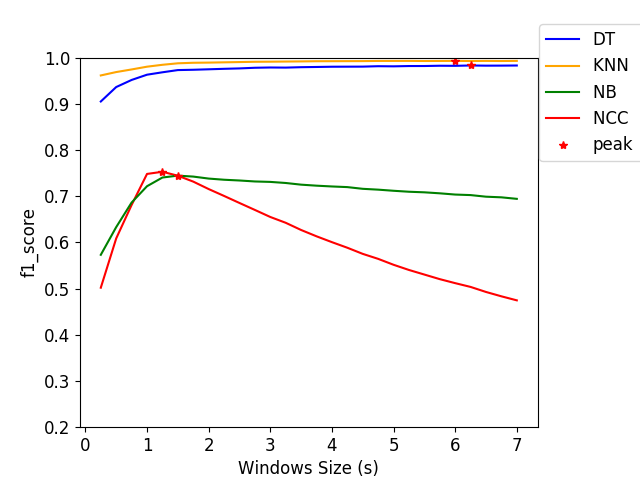}}
   
   \caption{Experiment 2 -- Overlapping windowing- k-fold CV process}
    \label{fig:O-iid-results}

\end{figure*}

\section{Discussion} \label{sec:discussion}

As can be seen by comparing Figures~\ref{fig:NO-iid-results} and~\ref{fig:NO-subjective-results}, using Subject CV instead of 
k-fold CV reduces the F1-score of KNN and DT by 10\% on average, which is substantial. It confirms our hypothesis that samples drawn from the 
same subject cannot be considered independent. In an ARP setup, k-fold CV overestimates the classification performance and should therefore be avoided.
   
The detrimental effect of k-fold CV is even larger when overlapping time windows are used. In this case, as can be seen by comparing Figures~\ref{fig:O-iid-results} and 
~\ref{fig:O-subjective-results}, Subject CV reduces the F1-score of KNN and DT by 16\% on average. This 
further confirms that within-subject dependencies between time windows account for a significant part of the performance measured through 
k-fold CV. Furthermore, for overlapping windows, the performance 
difference between k-fold CV and Subject CV increases with the window 
size. This is consistent with our previous comments, as the amount of 
overlap between overlapping windows, and therefore their correlation, 
also increases with the window size.

When using Subject CV, the impact of using overlapping windows is minor 
to negligible, as can be seen by comparing 
Figure~\ref{fig:NO-subjective-results} to 
Figure~\ref{fig:O-subjective-results}. This is in contradiction with 
the common argument that overlapping windows improve classification 
performance by bringing more data to the classifier. However, it also 
confirms our hypothesis that the performance increase coming from 
overlapping windows is in fact coming from the extra correlation 
between time windows, when k-fold CV is used.

The best observed F1-scores dropped from 0.96 in k-fold CV (reached 
for KNN and FS3) to 0.89 in Subject CV (KNN, FS3). This substantial 
difference opens an opportunity to improve the performance of 
HAR. We speculate from this study that building a dataset with a larger 
number of subjects would help reaching a better classification 
performance. Indeed, it seems difficult to capture the diversity of human behaviors in 33 
activities with only 17 subjects.

 As can be seen by comparing Figure~\ref{fig:NO-iid-results} and  Figure~\ref{fig:O-iid-results}, optimal window sizes are different when overlapping sliding windows are used and when they are not. This is mainly because overlapping windows increase the dependency between data points: as window sizes increase, the correlation between windows also increases. This does not happen with non-overlapping windows, which explains the difference in optimal window sizes.
 

The results in Figure~\ref{fig:NO-iid-results} are in agreement with the ones 
in~\cite{banos2014window}: F1-scores are not identical but close, 
and trends are generally consistent. This general agreement between our results 
and the ones in~\cite{banos2014window} reinforces our confidence in our 
results. The slight difference between our F1-score values might be 
coming from variations in the way the F1-scores are computed. F1-score 
is a harmonic mean between precision and recall; in case of multiclass 
targets, it can be computed in different ways. For instance, metrics 
can be calculated globally by counting total true positives, false 
negatives and false positives, or they can be calculated within each 
class. We calculated the F1-score globally, using the so-called ``micro" 
method in scikit-learn 0.20 \cite{pedregosa2011scikit}. The randomness 
involved in sample selection for k-fold CV, as well as other 
hyperparameters of the classifiers such as "classification criteria" and "max depth" for DT, might also explain 
some of the observed variations. In the future, we recommend to share all the parameters of the classifiers, for instance by sharing the scripts used for training, to improve reproducibility.

The size of segmented data by overlapping sliding windows technique is almost 9 times of data produced by non-overlapping one. As a result, the training time for classifiers on overlapping windowed datasets is also much higher than for non-overlapping ones. In spite of such increase in size and computation, this technique does not improve the performance of the classifiers when used with Subject CV. 

Choosing the right validation framework also depends on the target 
application. Here, we assumed that the ARP system would be used on a 
different population of subjects than it was trained on, which we believe
is the main use case for such systems. However, in some cases, it might
be relevant to train and test the system on the same subject, for instance
when active learning is used. In such situations, Subject CV would
obviously underestimate classification performance. Other forms of cross validation should be investigated for such situations. 

There might also be situations where Subject CV would still 
overestimate performance, in case other confounding factors are present.
For instance, data acquisition conditions (site, protocol, sensors and 
their parametrization), subject demographics, or even pre-processing 
steps might also introduce spurious correlations between samples that 
would lead to performance overestimations. Our results only studied one 
aspect of the larger problem of results evaluation, which should remain 
a constant concern.

\section{Conclusion} \label{sec:conclusion}

We conclude that k-fold CV overestimates the performance of HAR systems by about 10\% (16\% when overlapping windows are used), and should therefore not be used in this context. Instead, Subject CV provides a performance evaluation framework closer to the goal of activity recognition systems: we recommend its use for performance evaluation and hyperparameter tuning in ARP. We also conclude that overlapping sliding windows aggravate the performance overestimation done by k-fold CV, but do not improve the performance measured by Subject CV. Their added-value in our context therefore seems limited.  
\section{Reproducibility} 
All the source codes for conducted experiments are available in our GitHub repository \footnote{\url{http://www.github.com/big-data-lab-team/paper-generalizability-window-size}}. It contains the scripts to segment the dataset~\cite{banos2012benchmark} for different window sizes, feature sets and sliding window techniques. There is also a script for training and testing all mentioned classifiers on windowed datasets. Finally it also contains code to reproduce all presented figures in this paper.

\section{Acknowledgments}

This work was funded by a Strategic Project Grant from the Natural Sciences and Engineering Research Council of Canada. We thank Martin Khannouz for his useful comments. We also thank the authors of~\cite{banos2014window} for making their dataset publicly available.
\bibliographystyle{unsrt}
\balance
\bibliographystyle{ACM-Reference-Format}
\bibliography{sigkddExp}

\end{document}